\documentclass[runningheads]{llncs}
\usepackage{graphicx}
\usepackage[linesnumbered, ruled, vlined]{algorithm2e}
\usepackage{xcolor}
\usepackage{hyperref}
\usepackage{fancyhdr}
\begin{document}
\title{Path Finding for the Coalition of Co-operative Agents Acting in the Environment with Destructible Obstacles }
\titlerunning{Path Finding for the Coalition of Co-operative Agents ...}
%
\author{Anton Andreychuk\inst{1, 2} \and
Konstantin Yakovlev\inst{1, 3}}
\authorrunning{Andreychuk A. and Yakovlev K.}
%
\institute{Federal Research Center ``Computer Science and Control'' of Russian Academy of Sciences, Moscow, Russia
 \and Peoples' Friendship University of Russia (RUDN University), Moscow, Russia
 \and National Research University Higher School of Economics (NRU HSE), Moscow, Russia\\
\email{andreychuk@mail.com, yakovlev@isa.ru}}
\maketitle              
\begin{abstract}
The problem of planning a set of paths for the coalition of robots (agents) with different capabilities is considered in the paper. Some agents can modify the environment by destructing the obstacles thus allowing the other ones to shorten their paths to the goal. As a result the mutual solution of lower cost, e.g. time to completion, may be acquired. We suggest an original procedure to identify the obstacles for further removal that can be embedded into almost any heuristic search planner (we use Theta*) and evaluate it empirically. Results of the evaluation show that time-to-complete the mission can be decreased up to 9-12 \% by utilizing the proposed technique.

\end{abstract}
\section{Introduction}

Path planning for a point robot is usually considered in Artificial Intelligence and robotics as a task of finding a path on the graph whose nodes correspond to the positions the robot (agent) can occupy, and edges -- possible transitions between them. Voronoi diagrams \cite{lavrenov2017}, visibility graphs \cite{lozano1979}, grids \cite{yap2002} are the most widespread graphs used for path finding, with grids being the most simple and easy-to-construct discretizations of the workspace. To find a path on a grid typically one of the algorithms from A* family is used. A* \cite{hart1968} is a heuristic search algorithm that searches in state-space comprised of the elements (nodes) corresponding to certain graph vertices (grid cells or corners). There exist various modifications of A* that are suitable for grid-based path finding. In this work we utilize so-called any-angle path finders that do not constrain agent's moves to cardinal and diagonal ones only but rather allow to move into arbitrary direction as long as the endpoints of the move are tied to distinct grid elements. Among such algorithms Theta* \cite{daniel2010}, Field D* \cite{ferguson2006}, Anya \cite{harabor2016}, etc., can be named.

Abovementioned algorithms can not be directly applied to multi-robot path planning which is gaining more and more attention nowadays due to numerous applications in transport \cite{morris2016}, logistics \cite{wurman2008}, agriculture \cite{vu2017}, military \cite{khachumov2017} and other domains, but they can be modified to become base blocks of multi-agent path finders such as CBS \cite{sharon2015}, M* \cite{wagner2011}, MAPP \cite{wang2011}, \cite{andreychuk2017}, AA-SIPP(m) \cite{yakovlev2017}, etc. Typically those planners consider the interaction between the robots only spatial-wise by taking into account possible collisions and avoiding them.

In this work we investigate the case when robots can interact and co-operate by performing not only move-or-wait actions but modify-the-environment actions as well. This is similar to integration of task and motion planning \cite{kaelbling2011}, but unlike other researchers in this field we do not concentrate on task planning with grasping the objects, which is a typical scenario, but rather on task planning with path finding. The approach we suggest can be of particular interest to solving so-called smart relocation tasks \cite{panov2017}, \cite{panov2016} when the mission can not be accomplished without the robots helping each other. 

\section{Problem statement}
\begin{figure}
    \centering
    \includegraphics[width=0.9\textwidth]{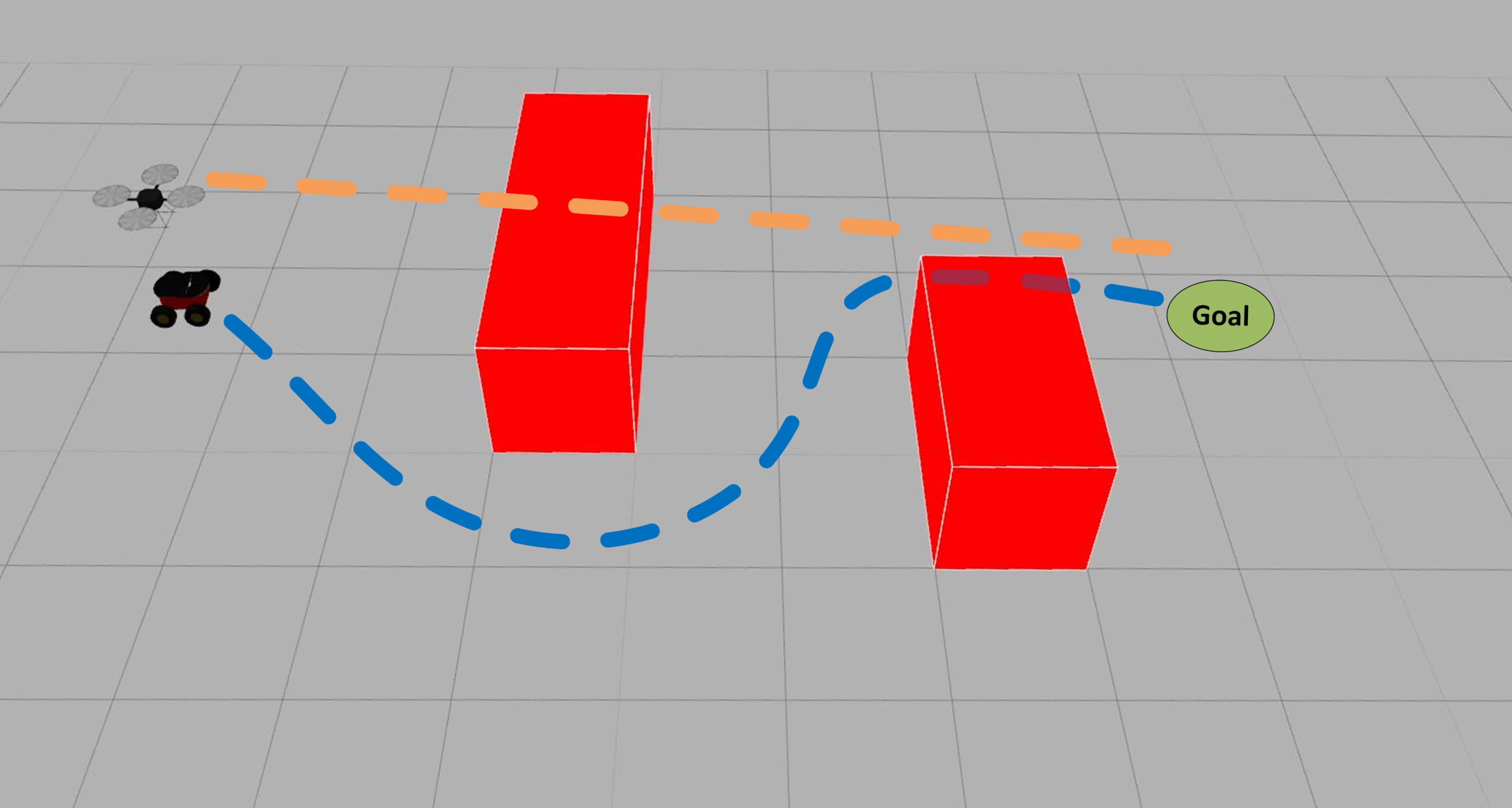}
    \caption{Heterogeneous group of mobile robots in a grid-world. Locations of the robots are tied to the centers of un-blocked grid cells. Flying robot can move directly to the goal flying above the obstacles, while wheeled robot has to circumnavigate them.}
    \label{fig1}
\end{figure}

Consider a coalition of heterogeneous robots that need to reach their respective goals in static, a-priory known environment, represented as a grid, composed of blocked and un-blocked cells. Without loss of generality we examine the case when only two heterogeneous robots, e.g. an UAV and a wheeled robot, are considered. The UAV can move directly to its goal, e.g. fly above all the obstacles, while the wheeled robot must circumnavigate them. An example of a modeled scenario is presented in \figurename \ref{fig1}.

The robots are different not only in the way they move, but they also may perform different set of actions. Wheeled robot can perform only move actions while UAV can destroy obstacles as well (at no cost). In order to do so it must first approach them. The problem now is to obtain coordinated mission completion plan composed of two sub-plans: one per each robot.The cost of the individual plan is the time needed to traverse the planned path, which is proportional to its length, so, without loss of generality the individual cost is the length of the path. Two metrics to measure the overall cost are considered, e.g. the flowtime (the sum of path lengths) and the makespan (maximum over path lenghts). We are interested in getting such solutions that have lower cost compared to the case when path planning is conducted independently by the robots. For the rest of the paper we assume that x- and y-coordinates of the start and goals location of two robots are equal, e.g. the UAV is hovering above the wheeled robot and its (x, y) goal location is the same. We will refer to the wheeled robot as to the first agent and to the UAV as to the second one.

\section{Method}

To decrease the cost of the initial non-cooperative solution one need to a) identify the obstacles that force the second agent to deviate from the shortest possible, e.g. straight-line, path to its goal; b) modify the original straight-line path of the second agent in such a way that it approaches each identified obstacle (thus destroying it); c) re-plan a path for the first agent. Also a grid pre-processing is needed in order to assign the unique identifiers to all the obstacles in the environment. This is done trivially by traversing the grid cells one by one, and every time a blocked cell is found, all adjacent blocked ones are traversed and assigned with a unique identifier. 

After finding and identifying all the obstacles on the grid, the trajectory for the first agent is planned using the modified heuristic search algorithm Theta*, which pseudocode is given in Algorithm 1. Besides finding the path the algorithm identifies obstacles whose removal can potentially shorten the length of such path. Detailed description of Theta* can be found in \cite{daniel2010}. We reference the reader to this paper for details and now proceed with the description of the proposed modifications. One of such modifications is that an additional data structure, $obstacles$, is introduced (lines 3-4) that stores the number of times each obstacle was hit during the search. Main loop is similar to the original Theta*, e.g. on each step the most promising state is retrieved and its successors are generated. These successors correspond to moves from the current cells to the neighbouring grid ones. If the move is infeasible due to the target cell being blocked it is discarded (as in conventional heuristic search path planner), but we also count the number of such blocked cells in lines 11-13. Thus, when the algorithm terminates, one obtains not only the information about whether the path was found or not\footnote[1]{The path itself can be reconstructed by iteratively tracing  backpointers from goal vertex until start is reached.}, but also the information about how many times the path-finder attempted to move through the particular obstacles.

\SetKwProg{Fn}{Function}{}{}
\SetInd{0.6em}{0.6em}
\begin{algorithm}[ht]
\caption{Theta* with counting interfering obstacles}
$parent(s_{start}) := s_{start}$; $g(s_{start}) := 0$\;
$OPEN:=\{s_{start}\}$; $CLOSED:=\emptyset$\;
\color{blue}{}
\For{each obstacle on grid as $o$}
{
    obstacles($o$) := 0\;
}
\color{black}{}
\While{$OPEN\neq\emptyset$}
{
    $s$ := state with minimal $f$-value from $OPEN$\;
    remove $s$ from $OPEN$ and add to $CLOSED$\;
    
    \If{$s = goal$}
    {
         \color{blue}{return obstacles and "path found"}\;
    }
    \For{each state in neighbours(s) as $s'$}
    {
        \color{blue}{\If{$s'$ is blocked}
        {
            obstacles(getObstacleAt($s'$))++\;
            continue\;
        }
        }
        \color{black}
        {
        \If{$s' \notin CLOSED$}
        {
            \If{$s' \notin OPEN$}
            {
                $g(s') := \infty$\;
            }
            updateVertex(s, s')\;
        }
        }
    }
}
 \color{blue}{return obstacles and "path not found"}\;
  \color{black}{}

\BlankLine
\Fn{updateVertex(s, s')}
{
    \If{lineOfSight(parent(s), s')}
    {
        $s := parent(s)$\;
    }
    \If{$g(s) + c(s,s') < g(s')$}
    {
        $g(s') := g(s) + c(s,s')$\;
        $f(s') := g(s') + h(s')$\;
        $parent(s') := s$\;
        insert/update $s'$ in $OPEN$\;
    }
}
\end{algorithm}

Obviously, if the vertices (cells) of the obstacle $o$ have never been considered during the search, e.g. $obstacles(o)$ is equal to 0, then this obstacle has no influence on robot's mission. If $obstacles(o)>0$ then removing $o$ might lead to a potentially shorter path. Thus, a simple criterion for removing an obstacle is suggested: $o=argmax_{o \in Obstacles}(obstacles(o))$, where $Obstacles$ stands for all obstacles on a grid. If it is possible to remove $n$ obstacles, then $n$ first ones with the largest values of $obstacles(o)$ are selected.

\begin{figure}[ht]
    \centering
    \includegraphics[width=0.9\textwidth]{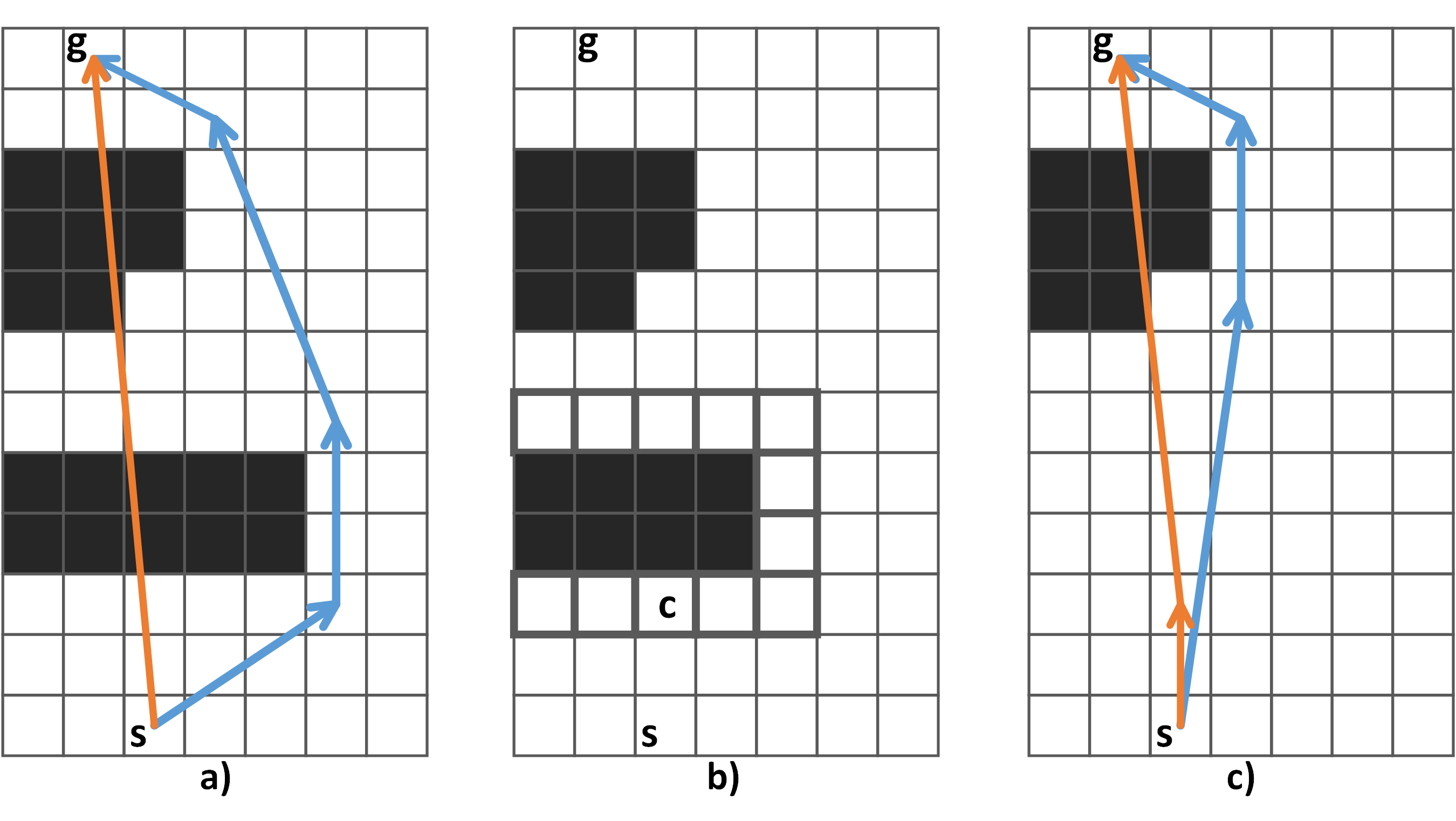}
    \caption{Robots' paths before and after removing the obstacle. a) initial paths for the ground robot and for the UAV (shown in blue and orange respectively) ; b) boundary of the obstacle to be removed (shown in bold) and the cell \textit{c} on this boundary that minimizes $dist(start, c) + dist (c, goal)$;  c) paths after removing the obstacle.}
    \label{fig2}
\end{figure}

After the obstacles are chosen, one needs to construct a trajectory for the second agent (flying robot), such that it passes through the vertices that are adjacent to the chosen obstacles. In case only one obstacle $o$ is going to be removed, the shortest path for the second agent can be found as follows. The distances from each cell comprising the boundary of $o$  to $start$ and $goal$ are calculated and such cell, $c$, is chosen that minimizes the distance $dist(start, c) + dist (c, goal)$, see \figurename \ref{fig2}b. This cell is used to form a path $[start, c, goal]$ . If more than one obstacle is going to be removed, such an approach becomes computationally burdensome. Instead, we suggest another procedure that does not guarantee finding the shortest path, but works much faster. 

Obviously, the shortest path from $start$ to $goal$ is the straight line segment connecting them $\langle start, goal\rangle$. Thus to minimize the length of the path that needs to pass through the vertices adjacent to the obstacles being removed, this path should be as close to this segment as possible. Therefore for each of the removing obstacles $o$ we look for such vertex (residing at the boundary of $o$) that minimizes the distance to $\langle start, goal\rangle$ segment. After all, the sought path for the second agent is constructed by aligning these vertices in order of increasing distance from $start$.

An example of removing an obstacle and re-planning paths for both agents is presented in \figurename \ref{fig2}. As one can see the path for the flying robot (marked in orange) is almost un-affected, while the path for the ground robot (marked in blue) becomes significantly shorter, which means that time-to-complete the mission lowers down.

\section{Experimental evaluation}

Described methods and algorithms were implemented in C++\footnote[2]{Source code is available at \url{https://github.com/PathPlanning/AStar-JPS-ThetaStar/tree/destroy_obs_and_replan}} and evaluated on a PC of the following configuration: OS - Windows 7, CPU - Intel Q8300 (2.5GHz), RAM - 2 GB. Descriptions of the real-world urban areas were extracted from OpenStreetMaps and used as the input. 100 maps each being ~1.2x1.2 km in size were transformed into 501x501 grids with blocked cells corresponding to buildings. Initially 200 instances per each grid were generated in such a way that the distance between the start and goal locations exceeded 960 meters (400 cells). Then the instances for which the length of the trajectory found by the original Theta* algorithm differed from the straight-line distance by no more than 10\% were discarded. Thus, a total of 1457 instances formed the resultant input.

To guide the search of the proposed modification of Theta* both un-weighted and weighted heurisic (Euclidean distance) was used, e.g. heuristic weight was set eiter to 1 (w = 1) or to 2 (w = 2). Using weighted heuristic makes the algorithm "greedy", i.e. more focused on the goal, as a result, it spends less time (and memory) to find a solution. The number of removed obstacles varied from 1 to 5.
The following performance indicators were tracked:

1) Path A -- path length of the first agent (i.e., of the agent that does not have the ability to modify the environment -- ground robot).

2) Number of nodes -- number of vertices that were processed and stored in memory in order to build path for the first agent. This indicator directly relates to memory consumption (the more vertices are stored the more memory is used).

3) Time -- runtime of the algorithm (excluding overheads, such as loading a map, saving a result, etc.).

4) Path B -- path length of the second agent (i.e., of the agent that has the ability to modify the environment -- flying robot).

These indicators were tracked both before and after the modification of the environment. The results of the conducted experiments are as follows.

\begin{figure}
    \centering
    \includegraphics[width=0.7\textwidth]{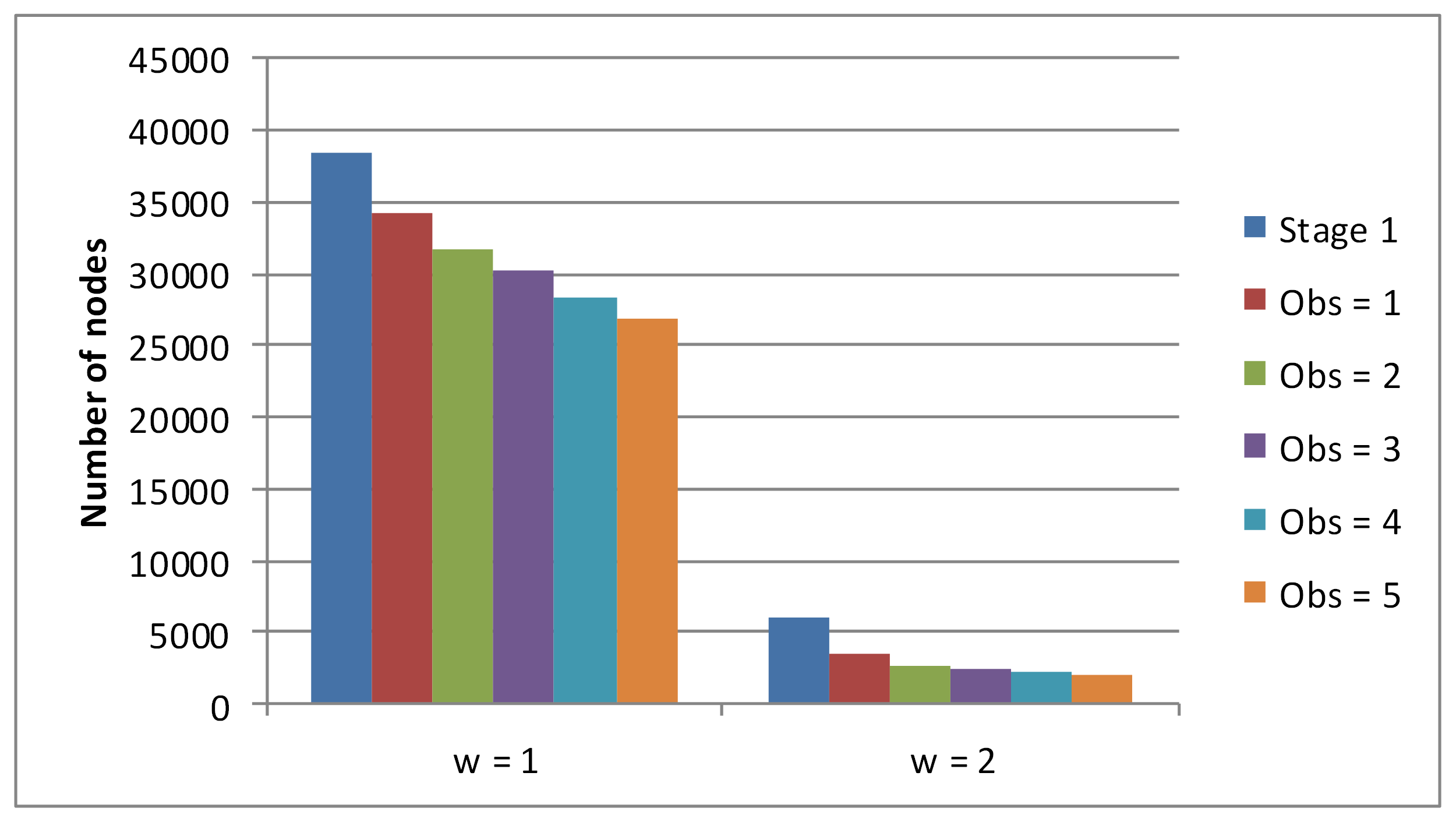}
    \caption{Memory consumption (number of processed nodes) before and after removing the obstacles.}
    \label{fig3}
\end{figure}

\figurename \ref{fig3} shows the average memory consumption after the first stage of planning (Stage 1), as well as after modifying the grid and removing the corresponding number of obstacles (Obs = 1, ..., Obs = 5). Left five columns correspond to the results obtained by the algorithm with un-weighted heuristic function (w = 1), while right 5 columns were gained with heuristic weight set to 2. One can note that removing obstacles leads to a notable reduction in the number of vertices processed by the path planning algorithm. If un-weighted heuristic is used (w = 1) than the memory consumption is reduced up to 30\%, moreover if the weighted heuristic function is utilized (w = 2) -- this consumption is reduced up to 65\%. 

\begin{figure}
    \centering
    \includegraphics[width=0.7\textwidth]{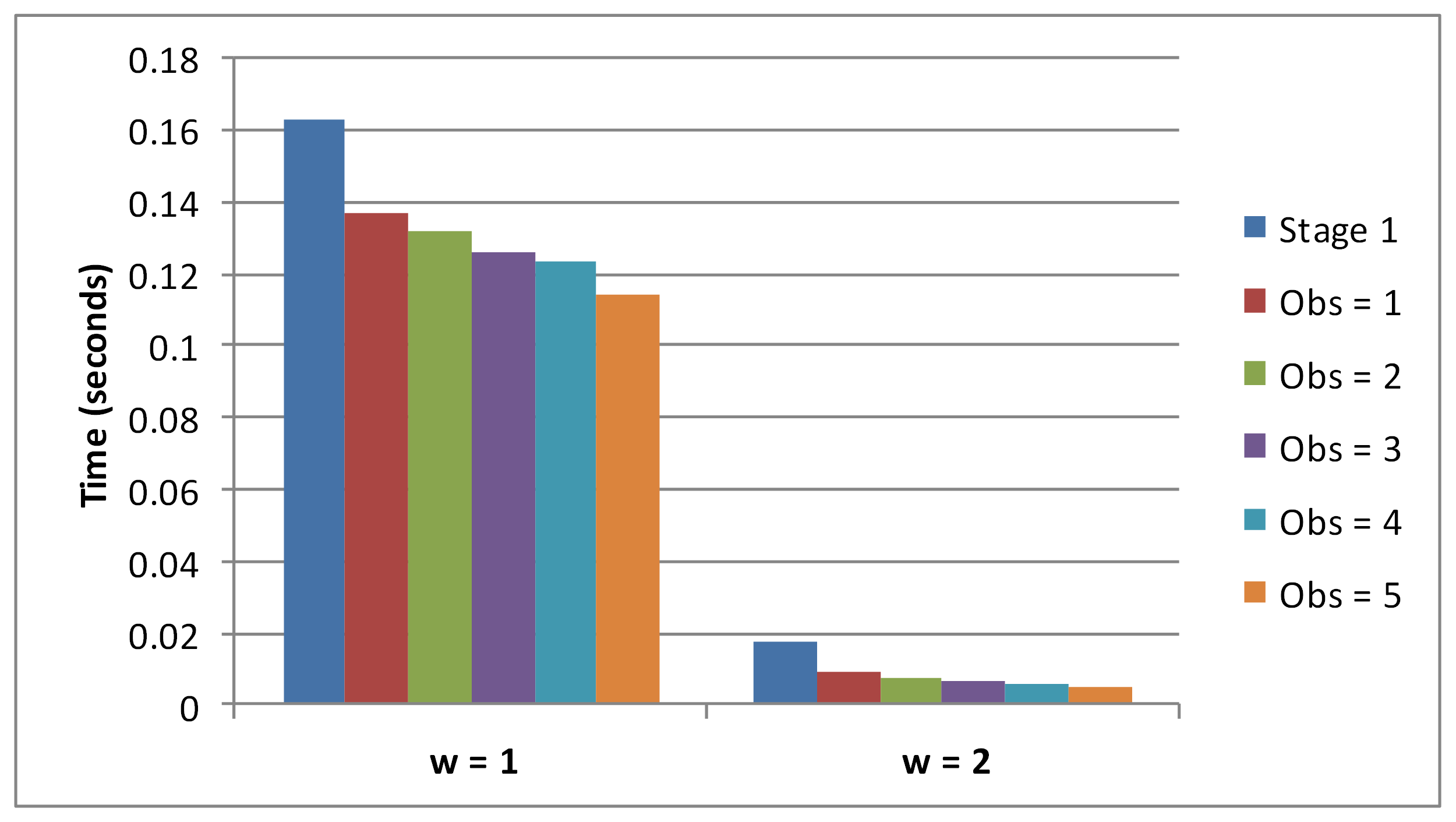}
    \caption{Planning time before and after removing the obstacles.}
    \label{fig4}
\end{figure}

Similar claims can be done w.r.t. runtime -- see \figurename \ref{fig4} for details. This figure shows the average amount of time which the algorithm spends to find the trajectories for both agents. Similarly to memory consumption, the runtime decreases when the number of removed obstacles increases. For w = 1 the runtime is reduced up to 29.8\%, and for w = 2 -- up to 71\%. 

To evaluate the quality of obtained solutions the following metrics were used:

- flowtime also known as sum-of-costs (SoC) -- the sum of the lengths of individual trajectories;

- makespan -- the maximum length of an individual trajectory.

Assuming that both agents move with identical speeds, the first metrics reflects the aggregate time-cost, associated with the mission; the second one shows when the last robot reaches its goal, so it can be seen as time to complete a mission.

\begin{figure}[ht]
    \centering
    \includegraphics[width=0.7\textwidth]{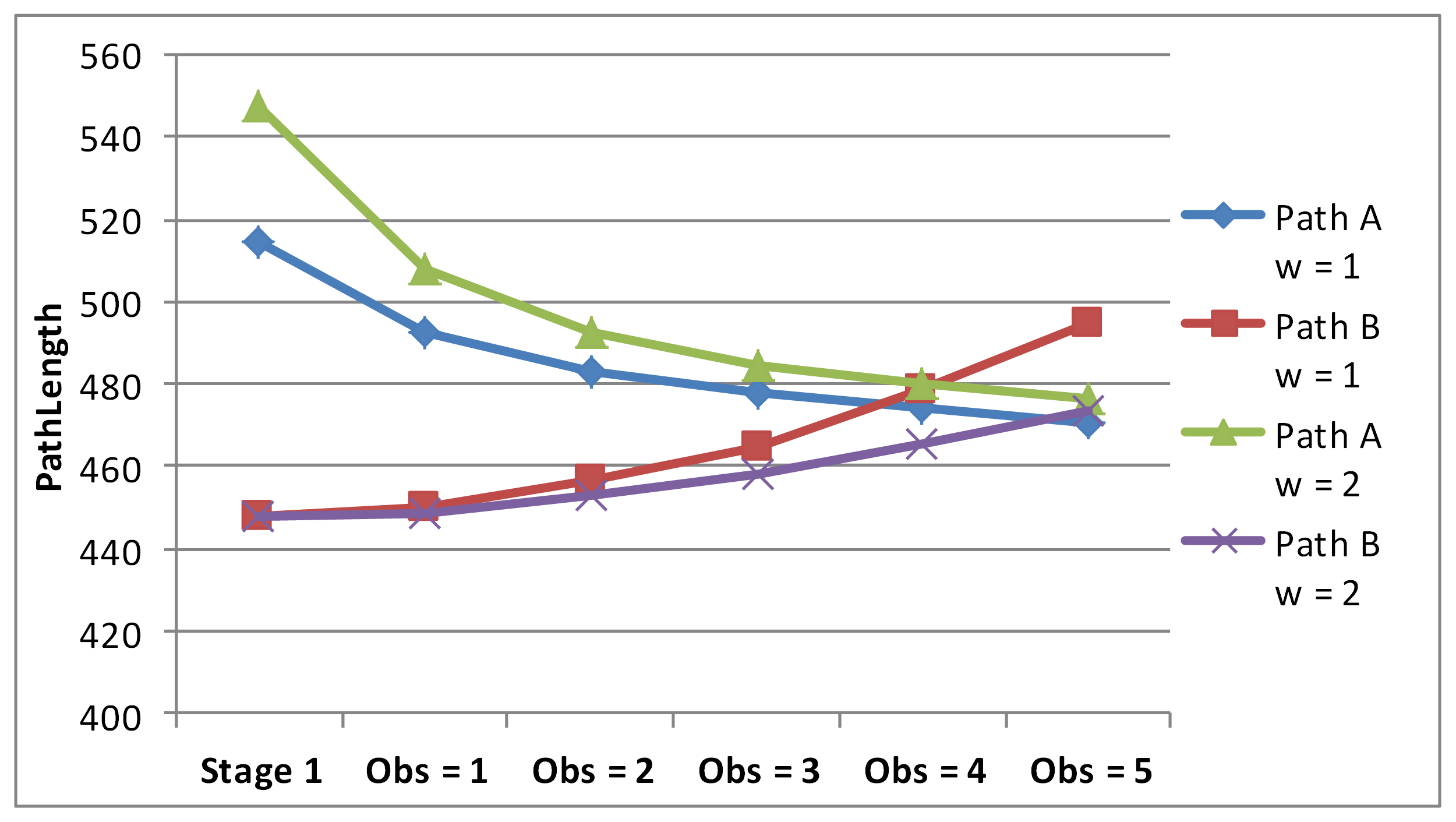}
    \caption{Path lengths of both agents before and after removing the obstacles.}
    \label{fig5}
\end{figure}

Before analyzing SoC and makespan let's look at the path lengths (averages) of both agents before and after obstacle removal -- see \figurename \ref{fig5}.

As can be seen, obstacle removal positively affects the path length of the first agent, but negatively affects the path length of the second one. So it's natural to assume that the number of removed obstacles should not be nor low nor high if one wants to reduce SoC and/or makespan.

\begin{figure}[ht]
    \centering
    \includegraphics[width=0.7\textwidth]{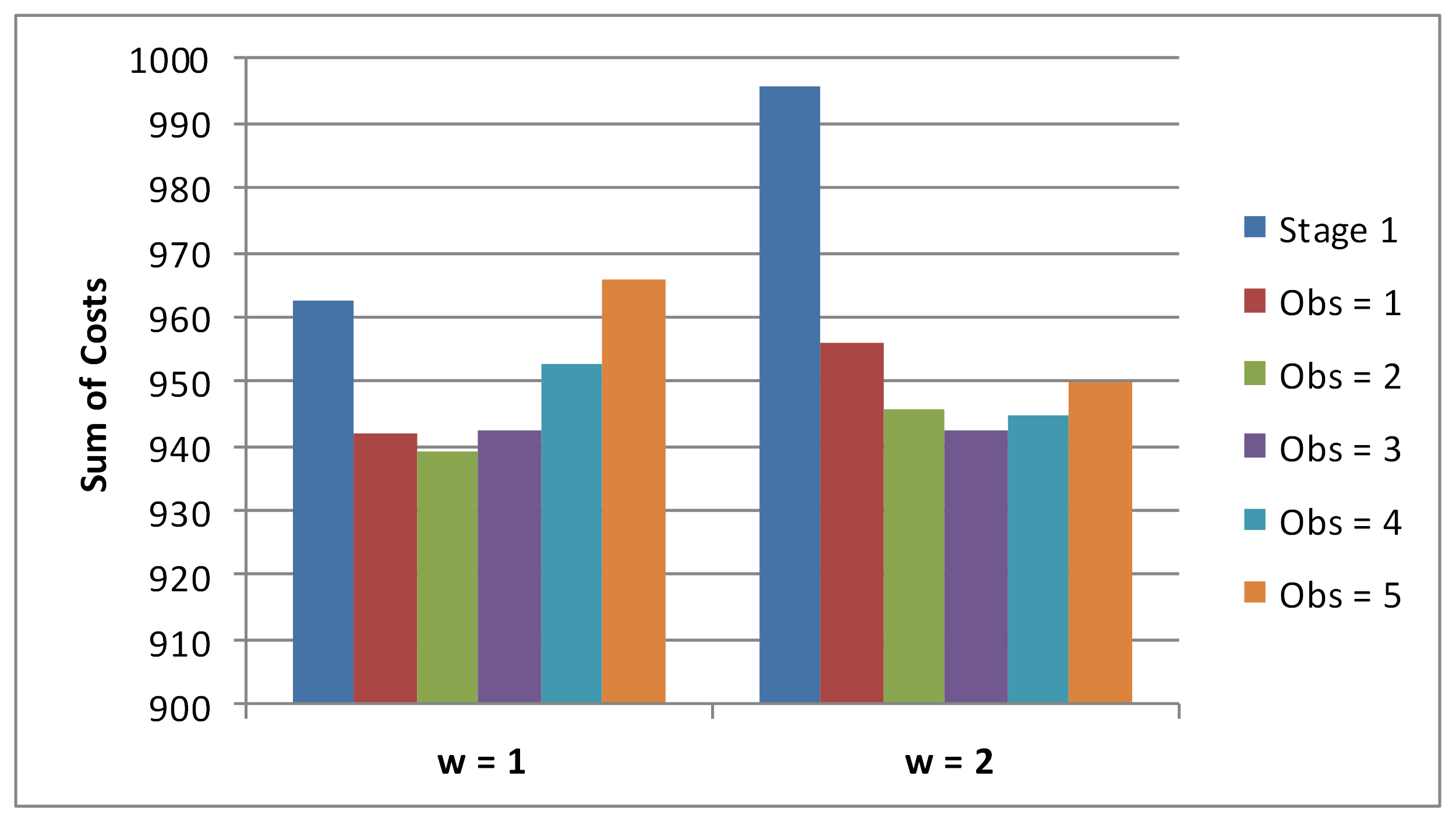}
    \caption{Sum-of-costs before and after removing the obstacles.}
    \label{fig6}
\end{figure}

Averaged SoC is depicted on \figurename \ref{fig6}. Analyzing this chart, one can claim that removing 2-3 obstacles leads to the best result, but percentage-wise the difference in SoC is not impressive (reduction by 3-5\%). This might be due to the input data and we believe that for other environments, e.g. maze-like environments or the ones with spiral-shaped obstacles, the reduction might be more notable.

\begin{figure}[ht]
    \centering
    \includegraphics[width=0.7\textwidth]{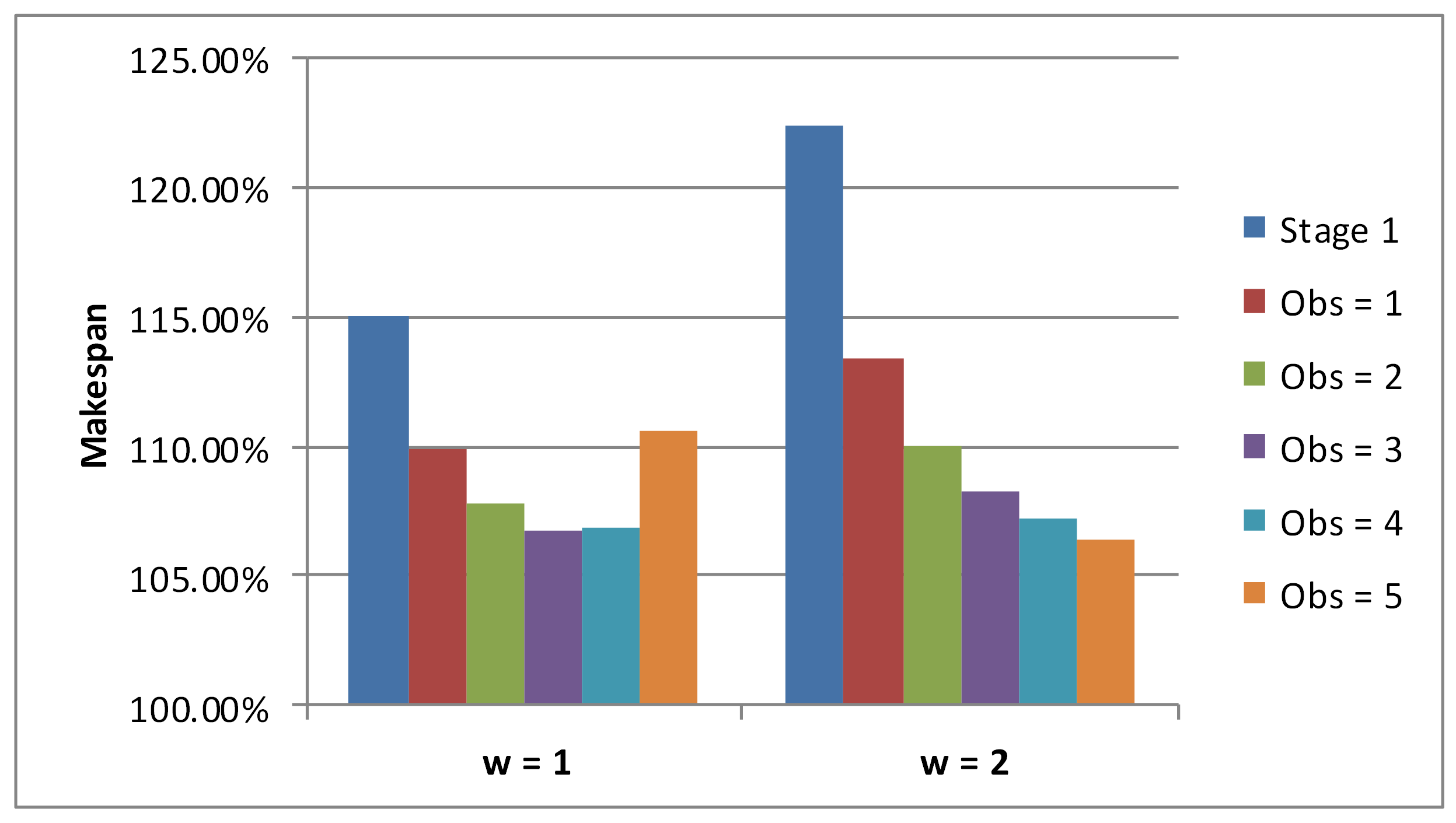}
    \caption{Normalized makespan before and after removing the obstacles.}
    \label{fig7}
\end{figure}

Average normalized values of makespan are shown on \figurename \ref{fig7}. We took the straight-line distance between the start and goal for 100\%. As one can see, removing 3-4 obstacles (5, if weighted heuristics is used) leads to the best performance. Time to complete the mission (e.g. makespan) reduces by 9-12\% is such cases, which is a notable reduction for numerous real-world applications.

\section{Conclusion}

We proposed an approach to plan a set of trajectories for the coalition of co-operative agents operating in the environment that can be modified by the actions of coalition members, e.g. some obstacles can be destroyed. The approach is based on the well-known heuristic search path planner, e.g. Theta*, as well as on a novel technique tailored to identify obstacles that obscure the path and thus should be potentially destroyed. Conducted experimental evaluation has shown that the suggested approach positively affects the solution quality, e.g. mission completion time (makespan) for the considered class of problems (navigation of ground and flying robots in urban environments).

\section{Acknowledgements}
This work was supported by the ``RUDN University Program 5-100'' (extracting data from OpenStreetMaps to conduct the experiments) and by the RSF project \#16-11-00048 (developing path planning methods and evaluating them).

%
%
%
 \bibliographystyle{splncs04}
 \bibliography{cites}

\end{document}